\documentclass[conference]{IEEEtran}
\IEEEoverridecommandlockouts
\usepackage{cite}
\usepackage{amsmath,amssymb,amsfonts}
\usepackage{algorithmic}
\usepackage{graphicx}
\usepackage{textcomp}
\usepackage{xcolor}
\usepackage{todonotes}
\usepackage{hyperref}
\usepackage{caption}
\usepackage{subcaption}
\usepackage{xspace}
\usepackage{tikz}
\usepackage{xintexpr, xinttools}
\usepackage{xcolor,colortbl}
\usepackage{flushend}

\makeatletter
\def\thickhline{%
  \noalign{\ifnum0=`}\fi\hrule \@height \thickarrayrulewidth \futurelet
   \reserved@a\@xthickhline}
\def\@xthickhline{\ifx\reserved@a\thickhline
               \vskip\doublerulesep
               \vskip-\thickarrayrulewidth
             \fi
      \ifnum0=`{\fi}}
\makeatother

\newlength{\thickarrayrulewidth}
\newcommand{\cTab}[1]
{%
    \xintifboolexpr {#1>=0}%
       {\cellcolor{green!\xinttheiexpr #1/0.8*100\relax}}
       {\cellcolor{magenta!\xinttheiexpr -#1/0.8*100\relax}}
    {#1}%
}%
\newcommand{\dTab}[1]
{%
    \xintifboolexpr {#1>=0}%
       {\cellcolor{green!\xinttheiexpr #1*600\relax}}
       {\cellcolor{magenta!\xinttheiexpr -#1*400\relax}}
    {#1}%
}
\def\BibTeX{{\rm B\kern-.05em{\sc i\kern-.025em b}\kern-.08em
    T\kern-.1667em\lower.7ex\hbox{E}\kern-.125emX}}
\hypersetup{colorlinks=true,linkcolor=blue,citecolor=black, urlcolor=blue}

\begin{document}

\title{Reducing numerical precision preserves classification accuracy in Mondrian Forests}
\author{Marc Vicuna$^1$, Martin Khannouz$^1$, Gregory Kiar$^2$, Yohan Chatelain$^1$, Tristan Glatard$^1$\\
    $^1$Department of Computer Science and Software Engineering,
    Concordia University, Montreal, Canada\\
    $^2$Center for the Developing Brain, Child Mind Institute, New York, NY, USA}
\maketitle

\begin{abstract}
    Mondrian Forests are a powerful data stream classification method, but their large memory footprint makes them ill-suited for low-resource platforms such as connected objects. We explored using reduced-precision floating-point representations to lower memory consumption and evaluated its effect on classification performance. We applied the Mondrian Forest implementation provided by OrpailleCC, a C++ collection of data stream algorithms, to two canonical datasets in human activity recognition: Recofit and Banos \emph{et al}. Results show that the precision of floating-point values used by tree nodes can be reduced from 64 bits to 8 bits with no significant difference in F1 score. In some cases, reduced precision was shown to improve classification performance, presumably due to its regularization effect. We conclude that numerical precision is a relevant hyperparameter in the  Mondrian Forest, and that commonly-used double precision values may not be necessary for optimal performance. Future work will evaluate the generalizability of these findings to other data stream classifiers.
\end{abstract}

\section{Introduction}
Mondrian Forests~\cite{lakshminarayanan2014mondrian}, an online variant of Random Forests, 
are powerful data stream classifiers that have been used in various applications. However, their considerable memory
footprint limits their applicability to low-memory devices such as connected objects, which are common in online
learning settings. We aim to decrease this footprint by reducing the numerical precision of floating-point data used throughout their training and application. Beginning from the OrpailleCC~\cite{OrpailleCC} implementation of Mondrian Forests, we applied this investigation to the popular use-case of human activity recognition using the Recofit~\cite{morris2014recofit} and Banos \emph{et al}~\cite{banos2012benchmark} datasets.

The floating-point data standard (IEEE-754)~\cite{microprocessor2019754} is universally supported across numerical computing. The popularity of this standard is due to its flexibility to store both very large and very small numbers
consistently. The size of these data structures, $\gamma$, is given by:
\begin{equation}
    \gamma=p+e+1,\label{eq}
\end{equation}
where $p$-bits are allocated to the mantissa (defining the value itself), $e$-bits are allocated to the exponent (its order of magnitude), and its sign fits in the remaining bit. Alongside the blueprint for constructing floating-point
data, the IEEE-754~\cite{microprocessor2019754} norm specifies floating-point formats commonly used in CPUs, such as binary32 ($e=8$, $p=23$) and binary64 ($e=11$, $p=52$), also known as single and double precision or ``floats'' and ``doubles''. Recently, deep learning has led to the emergence of reduced formats such as  Microsoft’s ms-fp8 Minifloats (8 bits), IEEE-754 binary16 (16 bits), bfloat16 (16 bits)~\cite{kalamkar2019study}, or posits (8, 16, 32, and 64 bits)~\cite{gustafson2017beating}. However, reduced formats implicitly store less information compared to double precision and should therefore be used carefully.

Mondrian classifiers are built upon like-named Mondrian processes~\cite{mondrianprocess}, which are multidimensional generalizations of Poisson processes. These can be interpreted as a recursive generative process that randomly makes univariate cuts partitioning the data space hierarchically into $k$-dimensional trees~\cite{bentley1975}. A key feature of these processes is that they are self consistent: a Mondrian process operating on the partition of some domain (e.g. a subdomain) is equivalent to a Mondrian process operating on the subdomain directly, allowing simultaneous subprocesses. While Mondrian processes can be infinite, Mondrian trees designate the classifiers formed by finite Mondrian processes, akin to decision trees. Each split is constructed by randomly selecting a feature and threshold, where the probability of selection is proportional to the normalized range of values for the feature. Finally, Mondrian Forests are constructed by ensembling a group of Mondrian trees, akin to Random Forest classifiers.

Data stream classifiers including Mondrian Forests,
Micro-Cluster Nearest Neighbours~\cite{microclus2017}, Hoeffding Trees~\cite{hoeffding}, and simple Feedforward Neural
Networks~\cite{feedforward1999} are all commonly used in online contexts and were previously compared 
in~\cite{khannouz2020benchmark} in the context of human activity recognition.  Results showed superior performance of 
Mondrian  Forests, and Hoeffding Trees. However, both methods remain too memory intensive for connected objects where memory is typically in the order of $\mathcal{O}$(100KB)~\cite{neblina}. Without compromising the breadth or depth of constructed forests, typically optimized hyperparameters which have a considerable impact on model performance, reducing the numerical precision used would potentially enable their application on memory-constrained platforms. However, given the inter-dependence of these parameters on model performance, numerical precision, memory consumption, 
and model parameterization must be studied together. This paper uniquely quantifies the impact of reducing numerical precision on the performance of Mondrian Forests classifiers for activity recognition and evaluates the impact these results have on their portability.

\section{Materials and Methods}

Using Verificarlo~\cite{denis2016verificarlo}, the numerical precision of Mondrian Forest classifiers as
implemented in OrpailleCC was controlled and modified. The reduced-precision models were tested with multiple human activity recognition datasets, Recofit and Banos \emph{et al.}, and performance was evaluated with F1 score.

\subsection{Mondrian Forest Implementation}

OrpailleCC's implementation of Mondrian Forests pre-allocates a fixed amount of memory shared by all the trees in the 
forest, leading to a constant memory footprint. When the amount of allocated memory is reached, tree growth is stopped 
in the forest.

Mondrian trees are tuned using three parameters: base count, discount factor, and budget. The base count is used to 
initialize the prediction in the root of the tree, the discount factor influences how reliant nodes
are upon their parent prediction, and the budget controls the tree depth. In our experiments, we use the same sets of hyperparameters as in~\cite{khannouz2020benchmark} (Table~\ref{tab1}).
\begin{table}[htbp]
    \caption{Hyperparameters used in the Mondrian Forests}
    \begin{center}
        \begin{tabular}{cccc}
            \thickhline
            \textbf{Number of trees} & \textbf{Base count} & \textbf{Discount} &
            \textbf{Budget}                                                          \\
            \hline
            \textbf{$1$}               & $0.0$                 & $1.0$               & $1.0$ \\
            \textbf{$5$}               & $0.0$                 & $1.0$               & $0.4$ \\
            \textbf{$10$}              & $0.0$                 & $1.0$               & $0.4$ \\
            \textbf{$50$}              & $0.0$                 & $1.0$               & $0.2$ \\
            \thickhline
        \end{tabular}
        \label{tab1}
    \end{center}
\end{table}

Under memory constraints, increasing the number of trees does not necessarily imply better performance as it
reduces tree depth. We tested Mondrian Forests with $1$, $5$, $10$ and $50$ trees, and in configurations with $0.6$~MB, 
$1.2$~MB or $3$~MB of memory allocated. 

\subsection{Simulating Reduced Precision} \label{intro:instr}
The experiments presented here rely on Verificarlo~\cite{denis2016verificarlo}, a tool that allows for the
perturbation and manipulation of floating-point data. Verificarlo is an LLVM-based compiler that instruments 
floating-point instructions through different backends. The VPREC~\cite{chatelain2019automatic} backend (for ``Virtual 
PRECision") was used here to simulate reduced floating-point formats. VPREC computes each instrumented ﬂoating-point 
operation using the original format (double precision in our case) and rounds the result to the desired precision, 
thereby ensuring correct rounding. VPREC also checks that the modified exponent is in the requested dynamic range $[e_{min}, e_{max}]$, where $e_{min}=2-2^{e-1}$ and $e_{max}=2^{e-1}-1$.

We compared two instrumentation approaches to reduce the precision of Mondrian Forests in OrpailleCC:

\paragraph{Node instrumentation (NI)}  reduces floating-point precision only in tree nodes bounds, since most of
the memory used by the forest is occupied by nodes. For the node instrumentation, new node bounds are rounded only
when they are stored. This approach reduces memory consumption at most by half, as in OrpailleCC the memory occupied
by nodes is equally divided between integers and floating-point numbers.

\paragraph{Whole instrumentation  (WI)} reduces precision in all floating-point values, i.e., node bounds, 
hyperparameters, and random splits. While whole instrumentation only marginally reduces memory consumption compared to 
node instrumentation,  it implies that most floating-point calculations could be done in a globally reduced floating-point 
format, leading to faster computations and lower energy consumption.

For both instrumentations, we tested Mondrian Forests on $52$ levels of precision, corresponding to
each precision ranging from the maximum for 64-bit floating-point data down to a single bit ($p$ in Equation~\ref{eq}). We also tested various exponent lengths ($e$ in Equation~\ref{eq}) ranging from $11$ to $2$. 

\subsection{Datasets}

We tested reduced precision on the two publicly-available datasets in human activity recognition. The Banos \emph{et al.} dataset~\cite{banos2012benchmark,banos2014dealing} (henceforth referred to as ``Banos'') is a human activity
dataset with $17$ participants and $9$ sensors per participant. Each sensor samples 3D acceleration, gyroscope,
and magnetic field, as well as orientation in a quaternion format, producing a total of $13$ values. Sensors
were sampled at $50$~Hz, and each sample was associated with one of $33$ possible activities. In addition to the
$33$ activities, an extra activity labelled $0$ indicated no specific activity. We preprocessed the Banos dataset
using non-overlapping  windows of one second ($50$ samples) and $6$ axes (acceleration and gyroscope of
the right forearm sensor). We computed the average and the standard deviation over the window as features for
each axis and assigned the most frequent label to the window. To ensure that our results were independent of the order of the data points in the dataset, $7$ random orderings were tested. 


The second dataset, Recofit~\cite{morris2014recofit}, is a human activity dataset containing $94$ participants.
Similar to the Banos dataset, the activity labeled $0$ indicated no specific activity and similar activities
were merged (see Table 2 in~\cite{behzad2019}). We preprocessed the dataset similarly to the Banos, using
non-overlapping windows of one second and $6$ axes of data (acceleration and gyroscope) from one sensor.
From these $6$ axes, the average and the standard deviation over the window were used as features. Previously
reported F1 scores were globally much lower in this dataset, which is hypothesized to be due to the higher number
of participants.

\section{Results}

\label{sec:results}

We compared the classification performance obtained with the node (NI) and whole (WI) instrumentations to the original
double precision Mondrian Forests implementation in OrpailleCC. F1 scores were evaluated using the prequential error, 
meaning that for each new data point the model was first tested and then trained. We computed the resulting F1 score 
every $50$ elements.

\subsection{Reduced precision minimally affects performance}

\begin{figure*}
     \centering
     \begin{subfigure}[htbp]{0.93\paperwidth}
       
        \centering
        \begin{tikzpicture}
        \draw (0, 0) node[inner sep=0] {\includegraphics[width=\textwidth]{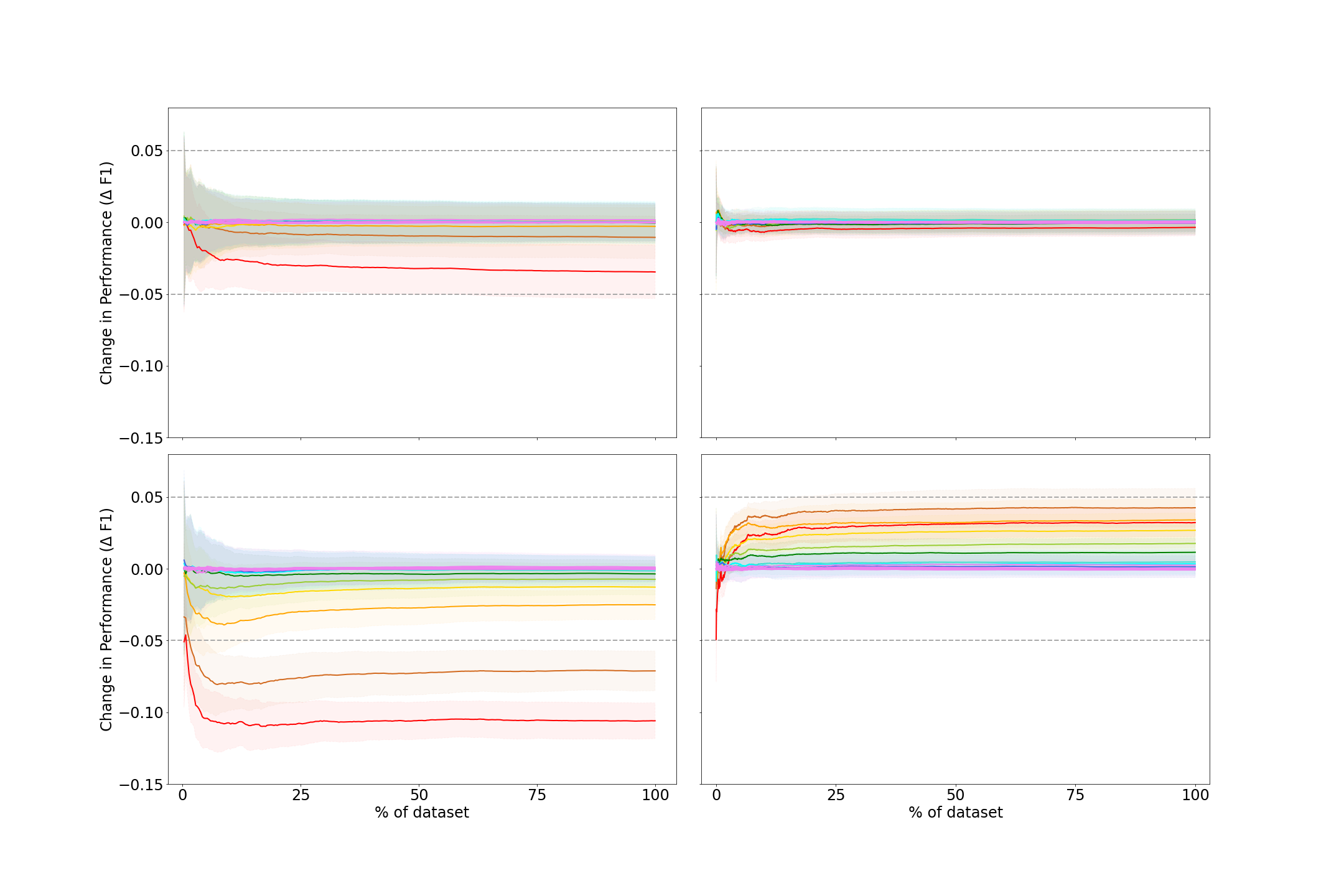}};
        
        \draw (-7.2, 0.5) node {(a)};
        \draw (0.8, 0.5) node {(b)};
        \draw (-7.2, -4.7) node {(c)};
        \draw (0.8, -4.7) node {(d)};
        \draw (-3.67, -5.8) node {Banos (14.2K elements)};
        \draw (4.35, -5.8) node {Recofit (84.8K elements)};
        \draw(0.2, -6.3) node {\textbf{Datasets}};
        \node[label=below:\label{difbn}\rotatebox{90}{Node}] at (-8.8,3.3) {};
        \node[label=below:\rotatebox{90}{Whole}] at (-8.8,-1.8) {};
        \node[label=below:\rotatebox{90}{\textbf{Instrumentations}}] at (-9.3,1.6) {};
        \end{tikzpicture}
         
    \end{subfigure}
     \hfill
     \begin{subfigure}[htbp]{0.55\paperwidth}
         \centering
         \includegraphics[width=\textwidth]{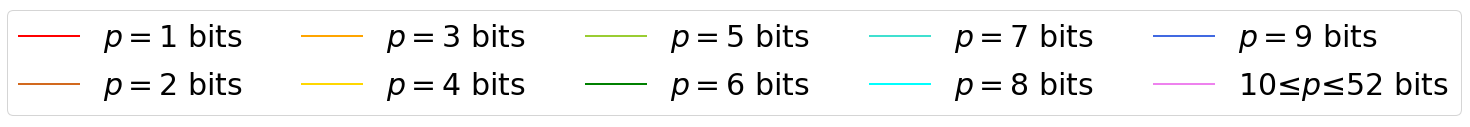}
     \end{subfigure}
    
    \caption{Difference between the F1 scores obtained at reduced and double precision ($p=52$ bits) with~$3.0$ MB of memory and $5$ trees.
    Negative values indicate that the instrumented implementation performed worse than the original. 
    Color shades represent the standard deviation in the corresponding Mondrian Forest.
    The grey lines at $-0.05$, $0.05$ are for reference only. The $7$ ordering of data points tested in Banos performed similarly.}
    \label{dif}
\end{figure*}
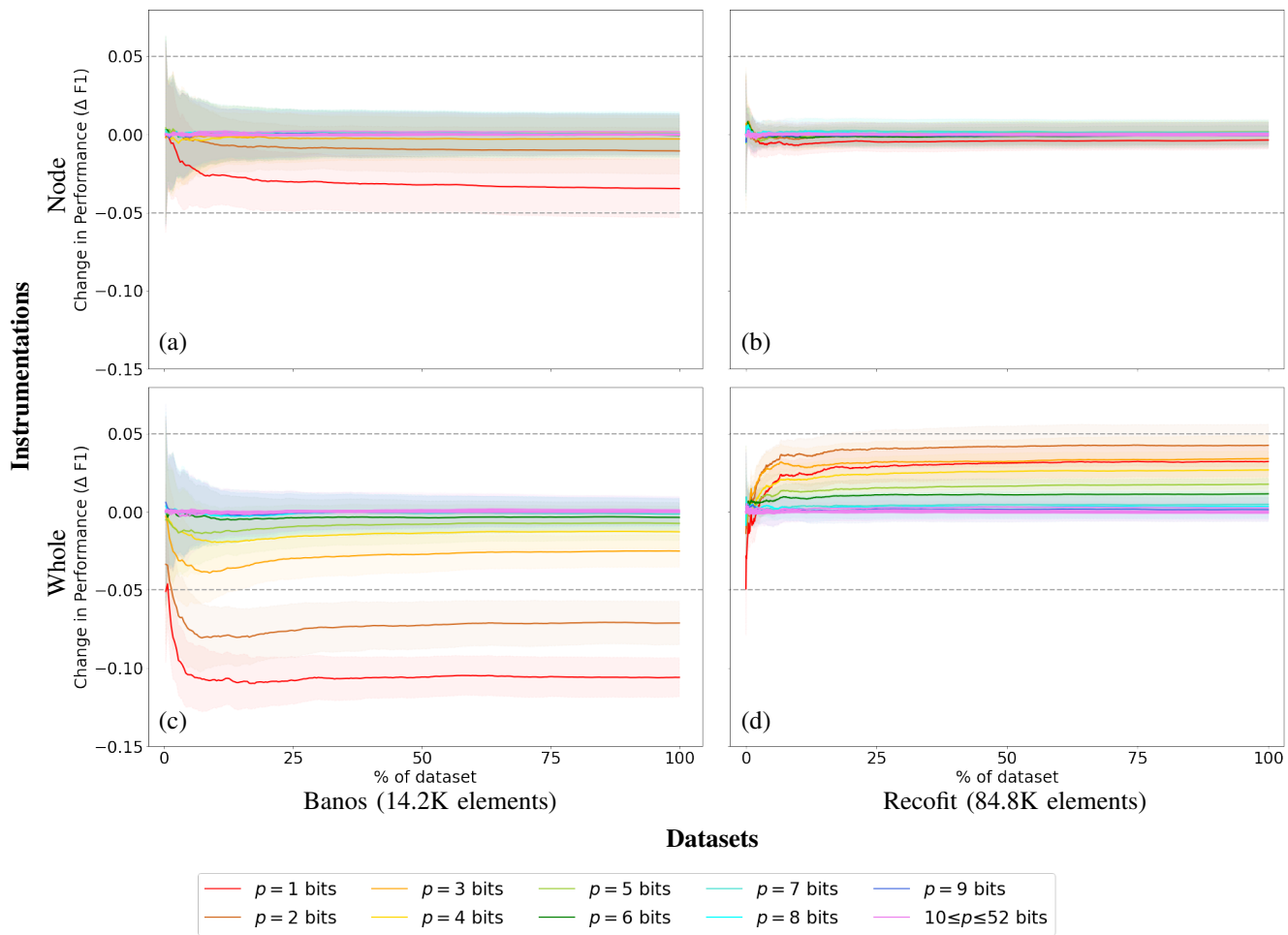

Figure~\ref{dif} shows the difference in F1 scores ($\Delta F1 = F1_{p=i} - F1_{p=52}$) between reduced-precision
and double precision classifiers across instrumentations and datasets. All datasets have been processed with an
exponent length of $11$ bits and with $3.0$~MB of memory.

Across all configurations, the F1 score difference stabilized within approximately the first $25\%$ of the dataset and remained stable beyond that point, indicating an overall good tolerance of Mondrian Forests to low precision.  

With node instrumentation, reductions in F1 scores remained within the standard deviation of performance for
configurations with $p \geq 2$  (Figures~\ref{dif}a and \ref{dif}b), indicating that using a low precision for the
node bounds has almost no influence on the F1 score. This can likely be attributed to the randomness of Mondrian 
Forests: given the randomness of the imposed splits, performance does not depend highly on their precision. Generally,
the error due to reduced precision remains within the existing randomness of node bounds.

In contrast, whole instrumentation (WI) shows significant deviations in F1 scores for configurations with
$p \leq 6$ (Figures \ref{dif}c and \ref{dif}d), indicating the high impact of precision loss in model 
hyperparameters. Distinctly, reducing the precision below $p = 6$ decreased the F1 score in the Banos dataset but 
increased the F1 score in the more complex Recofit dataset. 

Observed differences between NI and WI stem from low-precision representations of the split thresholds and budget,
since all other hyperparameters can be represented precisely with low precision (see Table \ref{tab1}). Low-precision split bounds and budget both lead to increased regularization. Low-precision split bounds reduce minimal distances between random splits, while a low-precision budget reduces tree depth. In complex datasets such as Recofit, which contain numerous activity classes and participants, regularization reduces overfitting. In contrast, simpler datasets such as Banos may not benefit from this form of regularization since the risk for overfitting is reduced.

\subsection{Reduced exponent size unimpactful down to $4$ bits}

Table~\ref{exponentTable} shows F1 score differences ($\Delta F1 = F1_{e=i} - F1_{e=11}$) between 
reduced-exponent and full-exponent Mondrian Forests. All classifiers used 52 bits of precision. The reported F1
score differences were calculated after having processed all the data points in the stream.

\begin{table}[htbp]
    \begin{minipage}{\linewidth}
        \caption{F1 score differences with reduced exponent length}
        \begin{center}
            \resizebox{\linewidth}{!}{
                \begin{tabular}{cccccccc}
                    \thickhline
                    
                    \textbf{Instru} & \textbf{Dataset} & \textbf{Memory} & \textbf{Trees} &\multicolumn{4}{c}{\textbf{Exponent length $e$}} \\
                    \cline{5-8} 
                    \multicolumn{4}{c}{} & \textbf{2 bits} & \textbf{3 bits} & \textbf{4 bits} & \textbf{5 bits}\\
                    \hline
                    NI         & Banos          & 0.6 MB        & 1           &\cTab{-0.4321}&\cTab{-0.0446}&\cTab{0.0000}&\cTab{0.0000}       \\
                    \cline{4-8} &               &              & 5           &\cTab{-0.5374}&\cTab{-0.0157}&\cTab{0.0000}&\cTab{0.0000}       \\
                    \cline{4-8} &               &              & 10          &\cTab{-0.5344}&\cTab{-0.0173}&\cTab{0.0000}&\cTab{0.0000}     \\
                    \cline{4-8} &               &              & 50          &\cTab{-0.3971}&\cTab{-0.0047}&\cTab{0.0000}&\cTab{0.0000}     \\
                    \cline{3-8} &               & 1.2 MB         & 1         &\cTab{-0.4696}&\cTab{-0.0904}&\cTab{0.0000}&\cTab{0.0000}  \\
                    \cline{4-8} &               &              & 5           &\cTab{-0.5937}&\cTab{-0.0275}&\cTab{0.0000}&\cTab{0.0000}      \\
                    \cline{4-8} &               &              & 10          &\cTab{-0.5805}&\cTab{-0.0217}&\cTab{0.0000}&\cTab{0.0000}     \\
                    \cline{4-8} &               &              & 50          &\cTab{-0.4771}&\cTab{-0.0045}&\cTab{0.0000}&\cTab{0.0000}      \\
                    \cline{3-8} &               & 3.0 MB         & 1         &\cTab{-0.4628}&\cTab{-0.0878}&\cTab{0.0000}&\cTab{0.0000}       \\
                    \cline{4-8} &               &              & 5           &\cTab{-0.6334}&\cTab{-0.0404}&\cTab{0.0000}&\cTab{0.0000}    \\
                    \cline{4-8} &               &              & 10          &\cTab{-0.6396}&\cTab{-0.0331}&\cTab{0.0000}&\cTab{0.0000}    \\
                    \cline{4-8} &               &              & 50          &\cTab{-0.5482}&\cTab{-0.0155}&\cTab{0.0000}&\cTab{0.0000}   \\
                    \cline{2-8} & Recofit          & 0.6 MB         & 1      &\cTab{-0.1763}&\cTab{-0.1291}&\cTab{-0.0031}&\cTab{0.0000}   \\
                    \cline{4-8} &               &              & 5           &\cTab{-0.1491}&\cTab{-0.1070}&\cTab{-0.0009}&\cTab{0.0000} \\
                    \cline{4-8} &               &              & 10          &\cTab{-0.0996}&\cTab{-0.0700}&\cTab{0.0000}&\cTab{0.0000}  \\
                    \cline{4-8} &               &              & 50          &\cTab{-0.0482}&\cTab{-0.0225}&\cTab{0.0000}&\cTab{0.0000}   \\
                    \cline{3-8} &               & 1.2 MB         & 1         &\cTab{-0.1948}&\cTab{-0.1470}&\cTab{-0.0038}&\cTab{0.0000}   \\
                    \cline{4-8} &               &              & 5           &\cTab{-0.1802}&\cTab{-0.1388}&\cTab{-0.0025}&\cTab{0.0000}  \\
                    \cline{4-8} &               &              & 10          &\cTab{-0.1501}&\cTab{-0.1090}&\cTab{-0.0009}&\cTab{0.0000}  \\
                    \cline{4-8} &               &              & 50          &\cTab{-0.0683}&\cTab{-0.0408}&\cTab{0.0000}&\cTab{0.0000}  \\
                    \cline{3-8} &               & 3.0 MB         & 1         &\cTab{-0.2127}&\cTab{-0.1601}&\cTab{-0.0042}&\cTab{0.0000} \\
                    \cline{4-8} &               &              & 5           &\cTab{-0.2214}&\cTab{-0.1726}&\cTab{-0.0018}&\cTab{0.0000}      \\
                    \cline{4-8} &               &              & 10          &\cTab{-0.1966}&\cTab{-0.1521}&\cTab{-0.0008}&\cTab{0.0000}    \\
                    \cline{4-8} &               &              & 50          &\cTab{-0.1003}&\cTab{-0.0717}&\cTab{-0.0001}&\cTab{0.0000}   \\

                    \hline
                    WI         & Banos          & 0.6 MB        & 1          &N/A&\cTab{-0.4628}&\cTab{-0.0006}&\cTab{0.0000}     \\
                    \cline{4-8} &               &              & 5           &N/A&\cTab{-0.5590}&\cTab{-0.0137}&\cTab{0.0000}   \\
                    \cline{4-8} &               &              & 10          &N/A&\cTab{-0.5404}&\cTab{-0.0309}&\cTab{0.0000}   \\
                    \cline{4-8} &               &              & 50          &N/A&\cTab{-0.3984}&\cTab{-0.1729}&\cTab{0.0000}   \\
                    \cline{3-8} &               & 1.2 MB         & 1         &N/A&\cTab{-0.4840}&\cTab{-0.0002}&\cTab{0.0000}  \\
                    \cline{4-8} &               &              & 5           &N/A&\cTab{-0.6130}&\cTab{-0.0009}&\cTab{0.0000}   \\
                    \cline{4-8} &               &              & 10          &N/A&\cTab{-0.5937}&\cTab{-0.0078}&\cTab{0.0000}    \\
                    \cline{4-8} &               &              & 50          &N/A&\cTab{-0.4798}&\cTab{-0.1351}&\cTab{0.0000}   \\
                    \cline{3-8} &               & 3.0 MB         & 1         &N/A&\cTab{-0.4923}&\cTab{-0.0001}&\cTab{0.0000}      \\
                    \cline{4-8} &               &              & 5           &N/A&\cTab{-0.6571}&\cTab{-0.0031}&\cTab{0.0000}   \\
                    \cline{4-8} &               &              & 10          &N/A&\cTab{-0.6586}&\cTab{-0.0091}&\cTab{0.0000}   \\
                    \cline{4-8} &               &              & 50          &N/A&\cTab{-0.5590}&\cTab{-0.1388}&\cTab{0.0000}  \\
                    \cline{2-8} & Recofit          & 0.6 MB         & 1      &N/A&\cTab{-0.1856}&\cTab{-0.0121}&\cTab{0.0000}   \\
                    \cline{4-8} &               &              & 5           &N/A&\cTab{-0.1552}&\cTab{-0.0389}&\cTab{0.0000} \\
                    \cline{4-8} &               &              & 10          &N/A&\cTab{-0.1069}&\cTab{-0.0260}&\cTab{0.0000}  \\
                    \cline{4-8} &               &              & 50          &N/A&\cTab{-0.0548}&\cTab{-0.0260}&\cTab{0.0000}   \\
                    \cline{3-8} &               & 1.2 MB         & 1         &N/A&\cTab{-0.2066}&\cTab{0.0088}&\cTab{0.0000}   \\
                    \cline{4-8} &               &              & 5           &N/A&\cTab{-0.1874}&\cTab{-0.0017}&\cTab{0.0000} \\
                    \cline{4-8} &               &              & 10          &N/A&\cTab{-0.1577}&\cTab{-0.0235}&\cTab{0.0000} \\
                    \cline{4-8} &               &              & 50          &N/A&\cTab{-0.0753}&\cTab{-0.0220}&\cTab{0.0000}  \\
                    \cline{3-8} &               & 3.0 MB         & 1         &N/A&\cTab{-0.2237}&\cTab{0.0159}&\cTab{0.0000} \\
                    \cline{4-8} &               &              & 5           &N/A&\cTab{-0.2321}&\cTab{0.0360}&\cTab{0.0000}      \\
                    \cline{4-8} &               &              & 10          &N/A&\cTab{-0.2058}&\cTab{0.0212}&\cTab{0.0000}    \\
                    \cline{4-8} &               &              & 50          &N/A&\cTab{-0.1073}&\cTab{-0.0107}&\cTab{0.0000}   \\
                    \thickhline
                \end{tabular}
            }
            \label{exponentTable}
        \end{center}
    \end{minipage}
\end{table}
For both datasets and instrumentations, all tree numbers, and all memory values, the F1 score did not significantly 
change until the exponent length was reduced to $\leq 3$ bits, with some exceptions in WI. For lower exponents lengths, forests with larger
numbers of trees tended to perform better. The whole instrumentation at $e=2$ was not evaluated since it causes segmentation faults due to the creation of infinite values.
For the tested datasets, an exponent size of 4 bits may be used without loss of classification performance. For other datasets, this limit could likely be estimated given awareness
of the dynamic range of the acquired signals.

\subsection{Larger forests are more robust to reductions in precision}

Table~\ref{precisionTable} shows F1 score differences between reduced and double precision for each set of tested 
hyperparameters. All classifiers used an exponent length of 11 bits. As before, the reported F1 score differences
were calculated after having processed all the complete stream of data.

\begin{table}[htbp]
    \begin{minipage}{\linewidth}
        \caption{F1 score differences with reduced  precision}
        \begin{center}
            \resizebox{\linewidth}{!}{
                \begin{tabular}{cccccccccc}
                    \thickhline
                    
                    \textbf{Instru} & \textbf{Dataset} & \textbf{Memory} & \textbf{Trees} &\multicolumn{6}{c}{\textbf{Precision $p$}} \\
                    \cline{5-10} 
                    \multicolumn{4}{c}{} & \textbf{1 bit} & \textbf{2 bits} & \textbf{3 bits} & \textbf{4 bits} & \textbf{5 bits} & \textbf{6 bits} \\
                    \hline
                    NI         & Banos          & 0.6 MB        & 1           &\dTab{-0.0514}&\dTab{-0.0132}&\dTab{0.0119}&\dTab{0.0146}&\dTab{0.0135}&\dTab{0.0134}       \\
                    \cline{4-10} &               &              & 5           &\dTab{-0.0032}&\dTab{0.0087}&\dTab{0.0098}&\dTab{-0.0065}&\dTab{0.0019}&\dTab{0.0007}    \\
                    \cline{4-10} &               &              & 10          &\dTab{-0.0070}&\dTab{-0.0011}&\dTab{-0.0014}&\dTab{0.0012}&\dTab{0.0007}&\dTab{-0.0022}   \\
                    \cline{4-10} &               &              & 50          &\dTab{-0.0077}&\dTab{-0.0026}&\dTab{-0.0040}&\dTab{-0.0029}&\dTab{-0.0006}&\dTab{-0.0015}  \\
                    \cline{3-10} &               & 1.2 MB         & 1         &\dTab{-0.0561}&\dTab{-0.0274}&\dTab{-0.0082}&\dTab{-0.0114}&\dTab{-0.0022}&\dTab{-0.0019}  \\
                    \cline{4-10} &               &              & 5           &\dTab{-0.0280}&\dTab{-0.0130}&\dTab{-0.0005}&\dTab{-0.0045}&\dTab{0.0013}&\dTab{-0.0025}   \\
                    \cline{4-10} &               &              & 10          &\dTab{-0.0152}&\dTab{-0.0030}&\dTab{-0.0049}&\dTab{0.0021}&\dTab{0.0040}&\dTab{-0.0029}  \\
                    \cline{4-10} &               &              & 50          &\dTab{-0.0036}&\dTab{0.0008}&\dTab{-0.0007}&\dTab{0.0008}&\dTab{0.0014}&\dTab{0.0027} \\
                    \cline{3-10} &               & 3.0 MB         & 1         &\dTab{-0.0692}&\dTab{-0.0323}&\dTab{-0.0058}&\dTab{-0.0070}&\dTab{0.0004}&\dTab{0.0006} \\
                    \cline{4-10} &               &              & 5           &\dTab{-0.0257}&\dTab{-0.0060}&\dTab{0.0037}&\dTab{-0.0022}&\dTab{0.0037}&\dTab{0.0017}  \\
                    \cline{4-10} &               &              & 10          &\dTab{-0.0155}&\dTab{0.0004}&\dTab{0.0001}&\dTab{0.0042}&\dTab{0.0014}&\dTab{0.0029}  \\
                    \cline{4-10} &               &              & 50          &\dTab{-0.0033}&\dTab{0.0022}&\dTab{-0.0006}&\dTab{-0.0010}&\dTab{0.0003}&\dTab{0.0012}   \\
                    \cline{2-10} & Recofit          & 0.6 MB         & 1      &\dTab{-0.0063}&\dTab{-0.0033}&\dTab{-0.0009}&\dTab{-0.0034}&\dTab{-0.0004}&\dTab{-0.0024}  \\
                    \cline{4-10} &               &              & 5           &\dTab{-0.0024}&\dTab{0.0003}&\dTab{0.0000}&\dTab{-0.0013}&\dTab{-0.0013}&\dTab{-0.0019}  \\
                    \cline{4-10} &               &              & 10          &\dTab{0.0006}&\dTab{0.0014}&\dTab{0.0005}&\dTab{0.0002}&\dTab{0.0012}&\dTab{0.0009}  \\
                    \cline{4-10} &               &              & 50          &\dTab{0.0018}&\dTab{0.0006}&\dTab{0.0005}&\dTab{0.0003}&\dTab{0.0008}&\dTab{0.0001}   \\
                    \cline{3-10} &               & 1.2 MB         & 1         &\dTab{-0.0154}&\dTab{-0.0051}&\dTab{-0.0012}&\dTab{-0.0014}&\dTab{-0.0025}&\dTab{-0.0012}  \\
                    \cline{4-10} &               &              & 5           &\dTab{-0.0031}&\dTab{-0.0011}&\dTab{-0.0002}&\dTab{-0.0004}&\dTab{-0.0007}&\dTab{0.0001}  \\
                    \cline{4-10} &               &              & 10          &\dTab{-0.0005}&\dTab{0.0025}&\dTab{0.0023}&\dTab{0.0010}&\dTab{0.0026}&\dTab{0.0010} \\
                    \cline{4-10} &               &              & 50          &\dTab{0.0018}&\dTab{-0.0006}&\dTab{0.0002}&\dTab{-0.0010}&\dTab{0.0001}&\dTab{-0.0005} \\
                    \cline{3-10} &               & 3.0 MB         & 1         &\dTab{-0.0140}&\dTab{-0.0072}&\dTab{-0.0007}&\dTab{-0.0005}&\dTab{-0.0005}&\dTab{-0.0003} \\
                    \cline{4-10} &               &              & 5           &\dTab{-0.0036}&\dTab{-0.0001}&\dTab{0.0018}&\dTab{-0.0001}&\dTab{0.0007}&\dTab{-0.0004}   \\
                    \cline{4-10} &               &              & 10          &\dTab{-0.0026}&\dTab{0.0019}&\dTab{0.0013}&\dTab{-0.0007}&\dTab{0.0012}&\dTab{0.0000}  \\
                    \cline{4-10} &               &              & 50          &\dTab{0.0007}&\dTab{0.0001}&\dTab{-0.0004}&\dTab{-0.0003}&\dTab{0.0004}&\dTab{-0.0011}  \\

                    \hline
                    WI        & Banos          & 0.6 MB         & 1          &\dTab{-0.0755}&\dTab{-0.0570}&\dTab{-0.0330}&\dTab{-0.0239}&\dTab{-0.0072}&\dTab{-0.0064}   \\
                    \cline{4-10} &               &              & 5          &\dTab{-0.0935}&\dTab{-0.0603}&\dTab{-0.0272}&\dTab{-0.0124}&\dTab{-0.0055}&\dTab{-0.0059}  \\
                    \cline{4-10} &               &              & 10         &\dTab{-0.1150}&\dTab{-0.0510}&\dTab{-0.0172}&\dTab{-0.0090}&\dTab{-0.0038}&\dTab{-0.0041}   \\
                    \cline{4-10} &               &              & 50         &\dTab{-0.2147}&\dTab{-0.1505}&\dTab{-0.0773}&\dTab{-0.0111}&\dTab{0.0020}&\dTab{0.0011}  \\
                    \cline{3-10} &               & 1.2 MB          & 1       &\dTab{-0.0705}&\dTab{-0.0647}&\dTab{-0.0271}&\dTab{-0.0193}&\dTab{-0.0165}&\dTab{0.0016}  \\
                    \cline{4-10} &               &              & 5          &\dTab{-0.1087}&\dTab{-0.0678}&\dTab{-0.0309}&\dTab{-0.0177}&\dTab{-0.0098}&\dTab{-0.0069} \\
                    \cline{4-10} &               &              & 10         &\dTab{-0.1203}&\dTab{-0.0624}&\dTab{-0.0217}&\dTab{-0.0107}&\dTab{-0.0011}&\dTab{-0.0030} \\
                    \cline{4-10} &               &              & 50         &\dTab{-0.2223}&\dTab{-0.1351}&\dTab{-0.0513}&\dTab{-0.0049}&\dTab{0.0025}&\dTab{0.0001}  \\
                    \cline{3-10} &               & 3.0 MB         & 1        &\dTab{-0.0737}&\dTab{-0.0660}&\dTab{-0.0331}&\dTab{-0.0348}&\dTab{-0.0127}&\dTab{-0.0051}  \\
                    \cline{4-10} &               &              & 5          &\dTab{-0.1038}&\dTab{-0.0729}&\dTab{-0.0294}&\dTab{-0.0149}&\dTab{-0.0106}&\dTab{-0.0068}   \\
                    \cline{4-10} &               &              & 10         &\dTab{-0.1228}&\dTab{-0.0648}&\dTab{-0.0199}&\dTab{-0.0122}&\dTab{-0.0033}&\dTab{-0.0042}   \\
                    \cline{4-10} &               &              & 50         &\dTab{-0.2035}&\dTab{-0.0994}&\dTab{-0.0405}&\dTab{-0.0079}&\dTab{0.0023}&\dTab{0.0002}  \\
                    \cline{2-10} & Recofit          & 0.6 MB         & 1     &\dTab{-0.0024}&\dTab{0.0121}&\dTab{0.0093}&\dTab{0.0058}&\dTab{0.0031}&\dTab{0.0014}  \\
                    \cline{4-10} &               &              & 5          &\dTab{0.0159}&\dTab{0.0214}&\dTab{0.0088}&\dTab{0.0076}&\dTab{0.0066}&\dTab{0.0026}   \\
                    \cline{4-10} &               &              & 10         &\dTab{0.0103}&\dTab{0.0206}&\dTab{0.0127}&\dTab{0.0123}&\dTab{0.0038}&\dTab{0.0024}  \\
                    \cline{4-10} &               &              & 50         &\dTab{0.0112}&\dTab{0.0077}&\dTab{0.0018}&\dTab{0.0019}&\dTab{0.0018}&\dTab{0.0007}   \\
                    \cline{3-10} &               & 1.2 MB         & 1        &\dTab{0.0084}&\dTab{0.0203}&\dTab{0.0099}&\dTab{0.0132}&\dTab{0.0046}&\dTab{0.0052}  \\
                    \cline{4-10} &               &              & 5          &\dTab{0.0197}&\dTab{0.0295}&\dTab{0.0176}&\dTab{0.0131}&\dTab{0.0043}&\dTab{0.0057}  \\
                    \cline{4-10} &               &              & 10         &\dTab{0.0091}&\dTab{0.0227}&\dTab{0.0135}&\dTab{0.0132}&\dTab{0.0052}&\dTab{0.0053}  \\
                    \cline{4-10} &               &              & 50         &\dTab{0.0071}&\dTab{0.0058}&\dTab{0.0010}&\dTab{0.0020}&\dTab{0.0021}&\dTab{0.0011}  \\
                    \cline{3-10} &               & 3.0 MB         & 1        &\dTab{0.0124}&\dTab{0.0282}&\dTab{0.0177}&\dTab{0.0148}&\dTab{0.0071}&\dTab{0.0105}   \\
                    \cline{4-10} &               &              & 5          &\dTab{0.0322}&\dTab{0.0426}&\dTab{0.0341}&\dTab{0.0268}&\dTab{0.0177}&\dTab{0.0115}  \\
                    \cline{4-10} &               &              & 10         &\dTab{0.0181}&\dTab{0.0356}&\dTab{0.0300}&\dTab{0.0264}&\dTab{0.0142}&\dTab{0.0060}  \\
                    \cline{4-10} &               &              & 50         &\dTab{0.0097}&\dTab{0.0147}&\dTab{0.0036}&\dTab{0.0038}&\dTab{0.0038}&\dTab{0.0004}  \\
                    \thickhline
                \end{tabular}
            }
            \label{precisionTable}
        \end{center}
    \end{minipage}
\end{table}

For NI and WI, classifiers with a higher number of trees were consistently less affected by the precision reduction.
This was likely because ensembling lowers the standard deviation in final predictions. The exception to the
above observation was when using the WI for prediction of the Banos dataset with $p<4$, where the loss of useful
complexity in the node splits appeared amplified by the resulting lower number of nodes per tree.

We also observed the F1 score differences at $p>2$ were non-significantly related to memory. The fact that the
reduction in precision did not affect memory allocations differently suggests that these benefits could be
translated to higher or lower memory allocations than the ones tested.

As mentioned in Figure~\ref{dif} and observed here consistently, the NI performance was stable on every precision 
tested, while WI had different tendencies depending on the dataset. Computations based on Table~\ref{precisionTable}
revealed that $p=3$ was the lowest precision where all results are within $2$ standard deviations of results at full 
precision.

\subsection{Expected memory footprint reduction}

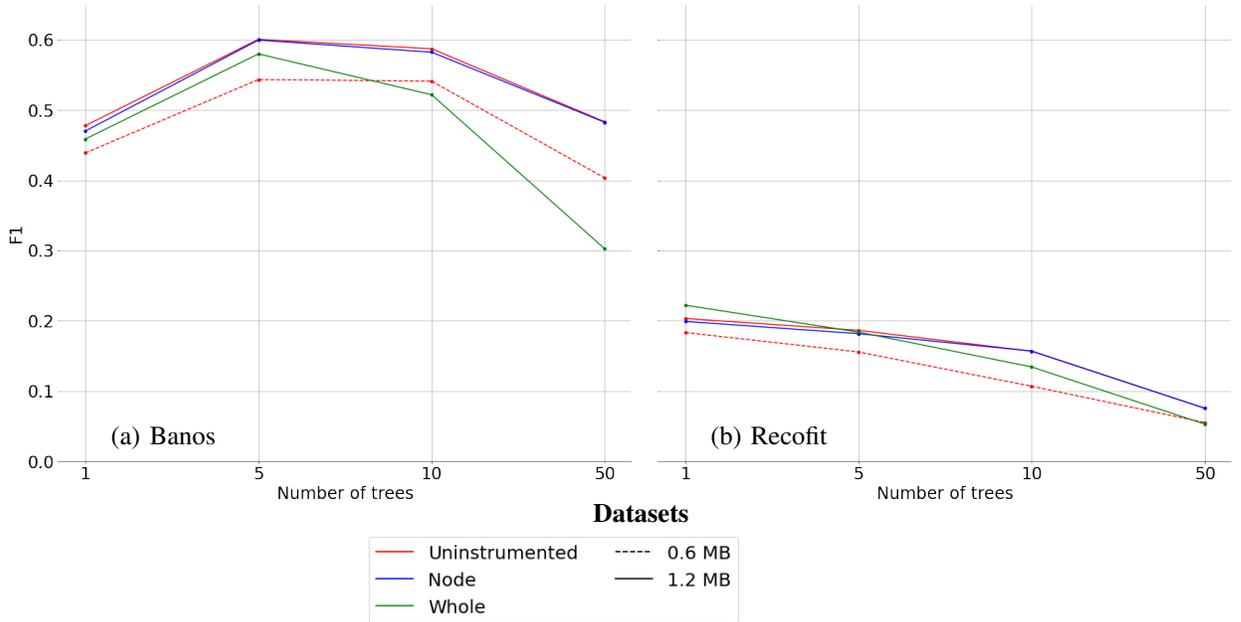
\begin{figure*}
     \centering
     \begin{subfigure}[htbp]{0.93\paperwidth}
       
        \centering
        \begin{tikzpicture}
        \draw (0, 0) node[inner sep=0] {\includegraphics[width=\textwidth]{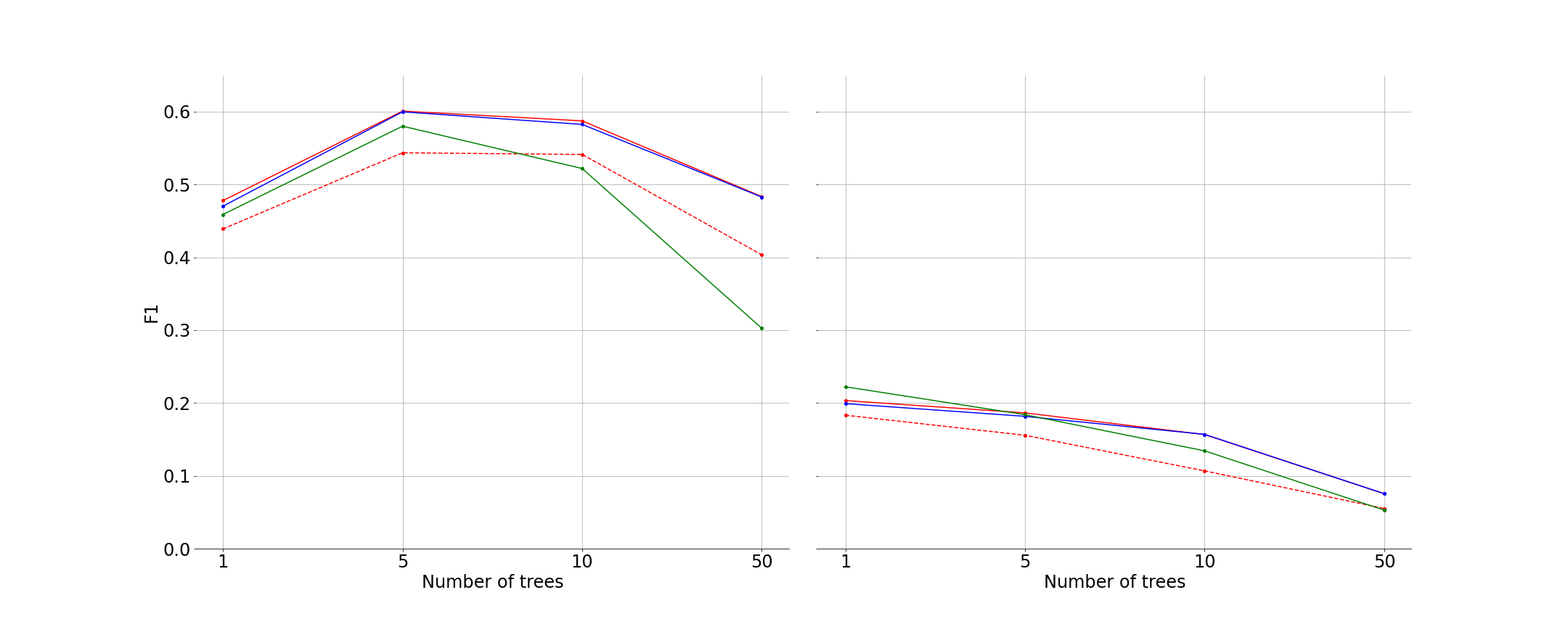}};
        \draw (-6.15, -2.7) node {(a) Banos};
        \draw (1.9, -2.7) node {(b) Recofit};
        \draw(0.2, -3.7) node {\textbf{Datasets}};
        \end{tikzpicture}
    \end{subfigure}
     \begin{subfigure}[htbp]{0.23\paperwidth}
         \centering
         \includegraphics[width=\textwidth]{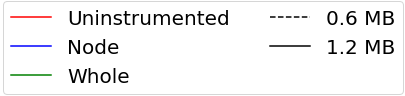}
     \end{subfigure}
        \caption{F1 Score of the uninstrumented classifiers with 0.6 MB and 1.2 MB,
        NI and WI classifiers at 4 bit exponent and 3 bit precision (8 bits) with 1.2 MB.}
        \label{gain}
\end{figure*}

Under constant memory constraints, reducing memory consumption could allow trees to grow deeper, thereby potentially 
improving classification performance. To quantify this performance improvement, Figure~\ref{gain} shows performance using double precision with $0.6$~MB of memory and reduced precision ($8$ bits) with $1.2$~MB of memory. As the execution platform  does not natively support 8-bit minifloats, we achieve similar memory scaling by doubling available memory. In reality, taking into account integer values in the implementation, the amount of memory consumed by the uninstrumented implementation is 
only $1.8 \times$ larger than for NI or WI. Indeed, the memory consumed by the uninstrumented implementation is $I + F$, 
where $I$ is the memory consumed by integer values and $F$ is the memory consumed by floating-point values ($I=F$ in our implementation), while the memory consumed by NI or WI with 8-bit minifloats is $I + F/8$. The limit of $0.6$~MB was selected as a realistic memory allocation bound
for connected objects~\cite{neblina}.

The F1 score obtained with NI at $1.2$~MB showed a $16.62\%$ average increase ($10.79\%$ on Banos,
$22.45\%$ on Recofit) compared to the uninstrumented classifier at $0.6$~MB, which quantifies the classification
performance improvement expected from reduced precision under memory constraints on platform-supporting minifloats. 
Besides, at 1.2~MB, NI has a very similar behavior to the uninstrumented 1.2~MB
classifier (superposed blue and red curves), confirming that 8-bit bounds for nodes
are practically possible and do not affect performance. 

As observed previously, WI was more variable, leading to inconsistent F1 score 
improvements or decreases depending on the number of trees.

\section{Related Work}


Mixed and reduced precision have been studied in various contexts with popular classifiers in recent years.
Rojek~\cite{gpurojek2019} dynamically reduced the energy consumption of Random Forest training up to $36\%$
by using mixed precision data. The objective of reducing energy consumption and memory on tree-based classifiers is 
common to our goal, however, Mondrian forests were not previously included in such a test. It is also not known
how tight the relationship is between energy savings obtained in the context of supercomputers versus those in
connected devices. Since Random Forests and Mondrian Forests are very similar, we hypothesize efficient
implementations of our improvements on Mondrian Forests would lead to lower energy consumption.

Yuval \emph{et al.}~\cite{yuvalstablerf2020} demonstrated the use of online and offline Random Forests to learn a 
parameterization from the coarse-grained output of a three-dimensional high-resolution idealized atmospheric
model. In their experiment, numerical precision is low (16 bits) and the classifiers are trained across resolutions. 
We note from their work that online random forests have more stable performance across resolutions than offline ones. This could indicate our observed performance stability across numerical precisions is correlated characteristics of online algorithms. 

Wang \emph{et al.}~\cite{8bitdeep2018} successfully reduced a wide spectrum of DNN data and computations
for training to 8-bit floating-point numbers with a potential of 2-4x improved throughput. 
Recent research~\cite{dynamicdeep2020} 
in Deep Neural Networks (DNNs) demonstrates that DNNs can have more than 99\% of their training 
reduced to half precision with similar accuracy to corresponding unreduced training.
The comparison of unreduced, dynamically reduced, and mostly reduced trained performances 
is similar to our approach, but deals with a very different machine learning model, and very
low-precision formats are not explored.

\section{Conclusion}

Node bounds in Mondrian Forests can be implemented
using an 8-bit floating-point format with negligible impact on the F1 score, reducing
the entire memory footprint by approximately $1.8 \times$ compared to double precision implementations
on 0.6 MB.  For a given amount of memory, decreasing node size allows Mondrian Forests to grow deeper trees 
or have more trees, potentially resulting in higher performance. In our configurations, performance 
improvements resulting from memory reduction at 0.6 MB improved the F1 score by 16.62\% on average. 
Improvements are hypothesized to be proportionally greater on lower memory at an optimal tree number. 

Reducing the precision of random cuts and the budget hyperparameter in Mondrian trees may improve or worsen classification performance depending on the dataset. While the observed performance improvements are encouraging, future research is required to fully understand when such improvements can be expected. In the meantime,
hyperparameterizing numerical precision using variable-length floating-points formats is a practical way to test this strategy.

Overall, reduced precision creates important opportunities for data stream classification in 
memory-constrained platforms such as connected objects. 
While we focused on reducing memory consumption, 
reduced precision would also decrease computation time substantially. Fully leveraging this result requires reduced-precision 
floating-point formats to be implemented in the target systems. Future research is
required to test this technique with other data stream classifiers.

\section{Code \& Data Availability} \label{repr}

The data, experimental results, plots, and code involved in our experiments are publicly available at \url{https://github.com/big-data-lab-team/benchmark-har-data-stream/tree/veripaille}.

\section*{Acknowledgments}
We thank Compute Canada and Calcul Québec for providing the computing infrastructure for our experiments. This work is partially funded by the Canada Research Chairs program, and by a Strategic Project Grant of the National Science and Engineering Research Council.

\bibliographystyle{ieeetr}
\bibliography{main.bbl}

\begin{thebibliography}{10}

\bibitem{lakshminarayanan2014mondrian}
B.~Lakshminarayanan, D.~M. Roy, and Y.~W. Teh, ``Mondrian forests: Efficient
  online random forests,'' {\em arXiv preprint arXiv:1406.2673}, 2014.

\bibitem{OrpailleCC}
M.~Khannouz, B.~Li, and T.~Glatard, ``{OrpailleCC}: a library for data stream
  analysis on embedded systems,'' in {\em Journal of Open Source Software},
  2019.

\bibitem{morris2014recofit}
D.~Morris, T.~S. Saponas, A.~Guillory, and I.~Kelner, ``Recofit: using a
  wearable sensor to find, recognize, and count repetitive exercises,'' in {\em
  Proceedings of the SIGCHI Conference on Human Factors in Computing Systems},
  pp.~3225--3234, 2014.

\bibitem{banos2012benchmark}
O.~Ba{\~n}os, M.~Damas, H.~Pomares, I.~Rojas, M.~A. T{\'o}th, and O.~Amft, ``A
  benchmark dataset to evaluate sensor displacement in activity recognition,''
  in {\em Proceedings of the 2012 ACM Conference on Ubiquitous Computing},
  pp.~1026--1035, 2012.

\bibitem{microprocessor2019754}
M.~S. Committee {\em et~al.}, ``754-2019-ieee standard for floating-point
  arithmetic,'' 2019.

\bibitem{kalamkar2019study}
D.~Kalamkar, D.~Mudigere, N.~Mellempudi, D.~Das, K.~Banerjee, S.~Avancha, D.~T.
  Vooturi, N.~Jammalamadaka, J.~Huang, H.~Yuen, {\em et~al.}, ``A study of
  bfloat16 for deep learning training,'' {\em arXiv preprint arXiv:1905.12322},
  2019.

\bibitem{gustafson2017beating}
J.~L. Gustafson and I.~T. Yonemoto, ``Beating floating point at its own game:
  Posit arithmetic,'' {\em Supercomputing Frontiers and Innovations}, vol.~4,
  no.~2, pp.~71--86, 2017.

\bibitem{mondrianprocess}
D.~M. Roy and Y.~Teh, ``The {M}ondrian process,'' in {\em Advances in Neural
  Information Processing Systems} (D.~Koller, D.~Schuurmans, Y.~Bengio, and
  L.~Bottou, eds.), vol.~21, Curran Associates, Inc., 2009.

\bibitem{bentley1975}
J.~L. Bentley, ``Multidimensional binary search trees used for associative
  searching,'' {\em Commun. ACM}, vol.~18, p.~509–517, Sept. 1975.

\bibitem{microclus2017}
M.~Tennant, F.~Stahl, O.~Rana, and J.~B. Gomes, ``Scalable real-time
  classification of data streams with concept drift,'' {\em Future Generation
  Computer Systems}, vol.~75, pp.~187--199, 2017.

\bibitem{hoeffding}
P.~Domingos and G.~Hulten, ``Mining high-speed data streams,'' in {\em
  Proceedings of the Sixth ACM SIGKDD International Conference on Knowledge
  Discovery and Data Mining}, KDD '00, (New York, NY, USA), p.~71–80,
  Association for Computing Machinery, 2000.

\bibitem{feedforward1999}
T.~L. Fine, {\em Feedforward neural network methodology}.
\newblock Statistics for engineering and information science., Springer, 2021.

\bibitem{khannouz2020benchmark}
M.~Khannouz and T.~Glatard, ``A benchmark of data stream classification for
  human activity recognition on connected objects,'' {\em Sensors}, vol.~20,
  no.~22, p.~6486, 2020.

\bibitem{neblina}
M.~Research, ``Neblina datasheet: Neblina v2 module bluetooth® smart 9 axis
  motion tracking,'' 2017.

\bibitem{denis2016verificarlo}
C.~Denis, P.~D.~O. Castro, and E.~Petit, ``Verificarlo: Checking floating point
  accuracy through monte carlo arithmetic,'' in {\em 2016 IEEE 23nd Symposium
  on Computer Arithmetic (ARITH)}, pp.~55--62, IEEE, 2016.

\bibitem{chatelain2019automatic}
Y.~Chatelain, E.~Petit, P.~de~Oliveira~Castro, G.~Lartigue, and D.~Defour,
  ``Automatic exploration of reduced floating-point representations in
  iterative methods,'' in {\em European Conference on Parallel Processing},
  pp.~481--494, Springer, 2019.

\bibitem{banos2014dealing}
O.~Banos, M.~A. Toth, M.~Damas, H.~Pomares, and I.~Rojas, ``Dealing with the
  effects of sensor displacement in wearable activity recognition,'' {\em
  Sensors}, vol.~14, no.~6, pp.~9995--10023, 2014.

\bibitem{behzad2019}
A.~Dehghani, O.~Sarbishei, T.~Glatard, and E.~Shihab, ``{A Quantitative
  Comparison of Overlapping and Non-Overlapping Sliding Windows for Human
  Activity Recognition Using Inertial Sensors},'' {\em Sensors}, vol.~19,
  no.~22, p.~5026, 2019.

\bibitem{gpurojek2019}
K.~Rojek, ``Machine learning method for energy reduction by utilizing dynamic
  mixed precision on gpu-based supercomputers,'' {\em Concurrency and
  Computation: Practice and Experience}, vol.~31, no.~6, p.~e4644, 2019.
\newblock e4644 cpe.4644.

\bibitem{yuvalstablerf2020}
J.~Yuval and P.~A. O’Gorman, ``Stable machine-learning parameterization of
  subgrid processes for climate modeling at a range of resolutions,'' {\em
  Nature Communications}, vol.~11, no.~1, p.~3295, 2020.

\bibitem{8bitdeep2018}
N.~Wang, J.~Choi, D.~Brand, C.~Chen, and K.~Gopalakrishnan, ``Training deep
  neural networks with 8-bit floating point numbers,'' {\em CoRR},
  vol.~abs/1812.08011, 2018.

\bibitem{dynamicdeep2020}
J.~Osorio, A.~Armejach, E.~Petit, and M.~Casas, ``A dynamic approach to
  accelerate deep learning training,'' {\em ICLR 2020 Conference Blind
  Submission}, 2020.

\end{thebibliography}
\end{document}